**Utilizing Multi-Agent Reinforcement Learning with Encoder-Decoder Architecture Agents to Identify**

**Optimal Resection Location in Glioblastoma Multiforme Patients**


Krishna Arun[1], Moinak Bhattacharya[2], Paras Goel[3]

[1]Eastlake High School,[1]Harvard Undergraduate OpenBio Laboratory Student Research Institute

[2]Department of Biomedical Informatics, Stonybrook University, NY

[3]School of Engineering and Applied Science, Columbia University, NY




**Abstract**

Currently, there is a noticeable lack of AI in the medical field to support doctors in treating heterogenous brain tumors such as Glioblastoma Multiforme (GBM), the deadliest human cancer in the world with a five-year survival rate of just 5.1%. This project develops an AI system offering the only end-to-end solution by aiding doctors with both diagnosis and treatment planning. In the diagnosis phase, a sequential decision-making framework consisting of 4 classification models (Convolutional Neural Networks and Support Vector Machine) are used. Each model progressively classifies the patient's brain into increasingly specific categories, with the final step being named diagnosis. For treatment planning, an RL system consisting of 3 generative models is used. First, the resection model (diffusion model) analyzes the diagnosed GBM MRI and predicts a possible resection outcome. Second, the radiotherapy model (Spatio-Temporal Vision Transformer) generates an MRI of the brain's progression after a user-defined number of weeks. Third, the chemotherapy model (Diffusion Model) produces the post-treatment MRI. A survival rate calculator (Convolutional Neural Network) then checks if the generated post treatment MRI has a survival rate within 15% of the user defined target. If not, a feedback loop using proximal policy optimization iterates over this system until an optimal resection location is identified. When compared to existing solutions, this project found 3 key findings: (1) Using a sequential decision-making framework consisting of 4 small diagnostic models reduced computing costs by 22.28x, (2) Transformers regression capabilities decreased tumor progression inference time by 113 hours, and (3) Applying Augmentations resembling Real-life situations improved overall DICE scores by 2.9%. These results project to increase survival rates by 0.9%, potentially saving approximately 2,250 lives.

*Keywords*: Glioblastoma Multiforme, Convolutional Neural Networks, Support Vector Machine Reinforcement Learning, Proximal Policy Optimization,



**Introduction**

Glioblastoma Multiforme (GBM) is the deadliest human cancer and most common malignant brain tumor in adults, with a five-year survival rate of just 5.1% (Ostrom et al., 2018) and approximately 250,000 new cases reported each year (De Vleeschouwer, 2017). Current treatment strategies typically involve a combination of surgical tumor removal (resection), radiotherapy, and chemotherapy. However, despite advancements in these approaches, survival outcomes have remained largely stagnant (Stupp et al., 2005). One reason for this is that its physical characteristics often resemble those of other neurological disorders, leading to a misdiagnosis rate of up to 28% (Iorgulescu et al., 2019). These delays in accurate diagnosis can reduce a patient's survival rate by as much as 50% before treatment even begins (Otani et al., 2023). Tumor diagnosis involves progressively classifying tumors into increasingly specific categories, with the final step being an accurate and reliable diagnosis. State-of-the-art approaches such as Vision Transformer (ViT)-L/16 use 303,317,095 parameters (Dosovitskiy et al., 2020), which require computational resources costing approximately $102,420 (estimated via Azure Compute Calculator) (Gupta, 2024). Such high costs restrict the use of these advanced diagnostic tools primarily to metropolitan hospitals, further widening the gap in healthcare accessibility for rural and under-resourced regions (Microsoft Learn, 2024).

GBM cells grow and infiltrate healthy brain tissue at a rate of approximately 1.4% per day (Egger et al., 2013). Because of this rapid progression, doctors must frequently monitor and evaluate the tumor's size and shape using brain tumor segmentation, the process of outlining the tumor's boundaries in MRI scans to track its development. Across a patient's treatment cycle, typically 5 to 8 segmentations are required; however, this process is time-consuming, often taking up to 10 hours per segmentation (Egger et al., 2013). The inability to perform segmentation in real time makes it difficult to adapt treatments dynamically. To address this, deep learning models such as DeepMedic and BraTS-AutoSeg have been developed to automate segmentation using data from the BraTS challenge. While these



models achieve high performance on standardized, they are often trained on ideal imaging conditions. As a result, they struggle to generalize lower-quality MRI scans common in rural or under-resourced hospitals, limiting their clinical utility in those settings (Singh et al., 2022; Kamnitsas et al., 2017).

Computational solutions for visualizing tumor growth, such as Lattice Boltzmann–based frameworks and the Glioma Solver, have also been developed. However, they require extensive computation times of 31 and 255 hours, respectively (Lipkova et al., 2022). To reduce this computational burden, these models often rely on simplified assumptions, such as uniform tumor growth and constant diffusion rates (Ennis et al., 2022; Cerri et al., 2021). Yet, due to the rapid progression and extreme biological heterogeneity of GBM, they are either too inaccurate to support personalized treatment planning or are too computationally intensive and slow when more realistic modeling is attempted (Pabisz et al., 2024; Mehta et al., 2023). Further limiting their clinical relevance, these models simulate tumor progression only when no treatment is applied, failing to account for how the tumor might respond to interventions such as surgery, chemotherapy, or radiation ultimately making them clinically not useful (Ravi et al., 2023).

This study proposes the development of the first end-to-end system to better address the doctor's pain points by developing 8 AI models split into two phases: Diagnosis and Treatment Planning (Figure 1). The diagnosis phase leverages a sequential decision-making framework comprising a series of 5 lightweight classification models (each with fewer than 20 million parameters). Each model progressively narrows the diagnostic scope by filtering out unlikely disease categories based on the intermediate outputs and learned feature representations from the brain MRI input. If the diagnosis phase finds the patient does in fact have GBM, then the output of the diagnosis phase (Confirmed MRI of GBM patient) will be the input of the treatment planning phase.

The standard treatment cycle for GBM typically involves surgical resection, followed by radiotherapy and chemotherapy. In the treatment planning phase, the system employs image-to-image



translation models that generate a forecasted MRI image of the tumor's appearance after each of these treatment steps. This can enable a clearer visualization of tumor progression and its morphological changes throughout the treatment sequence. Atop the forecasting models, using a CNN, the system also measures the generated treatment plan's survival rate, and will iteratively loop using Proximal Policy Optimization, until it satisfies the doctor's desired target (Liu et al., 2023).

**Figure 1**

Input-Output-Process Pipeline

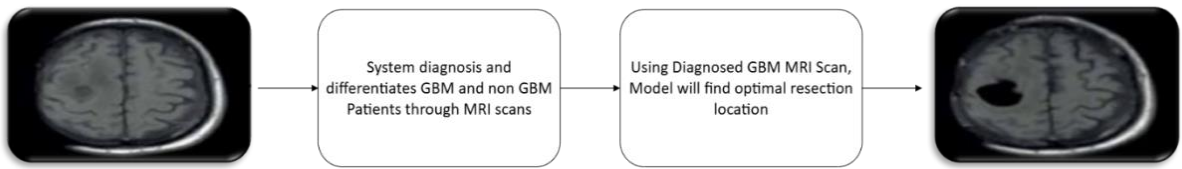

*End-to-end system pipeline for glioblastoma treatment planning. The process begins with a pre-treatment MRI, followed by a diagnostic system that distinguishes GBM from non-GBM cases. Upon diagnosis, a treatment planning model determines the optimal resection location, resulting in a simulated post-treatment MRI.*

## Contributions

This study presents key contributions in Medical Imaging and Oncological treatment using Machine Learning based frameworks:

a)  We developed a sequential decision-making framework consisting of a chain of multiple small parameter classification CNN models to predict patient diagnosis using the Pre Treatment MRI.

b)  We introduce a Spatio-temporal Vision Transformer with a novel decoder that models the effects of radiotherapy on tumor progression across multiple future time points.

c)  We enhance the preprocessing pipeline by introducing augmentations that resemble the medical imaging of less resourced hospitals to improve the system's ability to generalize to diverse clinical conditions.



## Methods

### Data Curation

To develop the proposed agentic AI framework for optimal treatment planning in GBM patients, a total of 6,560 patient cases comprising T1CE MRI scans and quantitative radiomic features were curated from seven reputable datasets housed in the Cancer Imaging Archive (Clark et al., 2013). This included datasets such as the BraTS dataset (Menze et al., 2015), the ReMIND dataset (Peng et al., 2021), and the Lumiere dataset (Zhou et al., 2020). Each dataset was developed by trusted medical institutions and was diverse in both patient demographics and tumor states. In most scans, the vast majority of voxels correspond to normal brain tissue, with only a small fraction representing tumor structures such as the enhancing core, edema, or necrotic center (Bakas et al., 2017). Because of this, the T1CE MRI imaging modality was primarily utilized due to its effectiveness in delineating tumor boundaries (Gordillo et al., 2013). Prior to data augmentation, the datasets underwent preprocessing. Duplicate entries, corrupted images, and incomplete cases were systematically removed to enhance data integrity (Tustison et al., 2014). All MRI scans were standardized to a uniform resolution of 64×64 pixels. Intensity normalization was then performed to adjust pixel values, ensuring consistency across imaging data (Nyúl & Udupa, 1999). Furthermore, Gaussian smoothing was applied to reduce image noise and enhance model learning efficiency (Buades et al., 2005).

To enhance model robustness and better simulate real-world clinical heterogeneity, particularly in under-resourced settings, the data augmentation pipeline incorporated domain-specific transformations in addition to conventional techniques (Shorten & Khoshgoftaar, 2019). Random bias field augmentation was employed to emulate intensity non-uniformities commonly introduced by magnetic field inhomogeneities and coil sensitivity artifacts inherent to MRI acquisition (Tustison et al., 2010). Elastic deformations were applied to simulate biologically plausible anatomical variability and intraoperative brain shifts, thereby capturing post-surgical morphological distortions (Simard,



Steinkraus, & Platt, 2003). Affine transformations, including scaling, translation, and rotation, were utilized to reflect inconsistencies in patient positioning across different imaging sessions (Perez & Wang, 2017). Additionally, random Gaussian noise was introduced to replicate distortions frequently observed in low-resolution or older MRI scanners, where hardware limitations can degrade image quality (Buades, Coll, & Morel, 2005). These medically informed augmentations were chosen to approximate the imaging variability observed in rural and heterogeneous clinical environments and expanded the dataset size by 10x. The augmented dataset was stratified into training and validation cohorts, with 80% allocated for training and 20% reserved for validation and performance benchmarking (Goodfellow, Bengio, & Courville, 2016).

**Phase 1: Diagnosis**

The diagnostic phase is structured as a sequential decision-making system (Figure 2) consisting of five classification models, each designed to progressively narrow the disease scope using an inputted MRI (Silver et al., 2021). This approach mimics a clinical decision-making pipeline, allowing each model to filter out unlikely diagnostic paths before passing its output to the next stage (Lundervold & Lundervold, 2019). This not only improves interpretability but also significantly reduces computational costs compared to using a single monolithic model (Rajpurkar et al., 2017).

**Figure 2**

Diagnosis Phase System Diagram

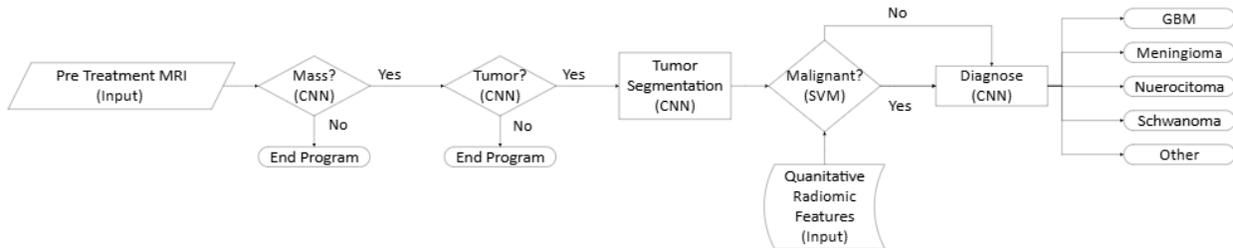

*Automated brain tumor diagnosis workflow using deep learning and machine learning approaches.*



From a high-level architectural perspective, the system operates as follows (Figure 2). The first CNN performs binary classification to determine whether the input MRI contains any abnormal mass (Krizhevsky, Sutskever, & Hinton, 2012). This model uses a shallow architecture (6 layers) to detect structural abnormalities without being biased toward any specific pathology for generalizability (LeCun, Bengio, & Hinton, 2015). If no mass is detected, the system halts with a "No Tumor" prediction. If a mass is detected, the image is passed to a second binary classification CNN which classifies whether the mass represents a tumor or a non-tumor mass (e.g., cyst) (Esteva et al., 2017). This model uses the same architecture as the first CNN architecture. Once a tumor is identified, an nnU-Net segmentation model, using automatic parameter tuning to handle the heavy class imbalance, segments the brain into edema, enhancing tumor, and tumor core regions (Isensee et al., 2021). Using the segmented MRI image, doctors must input 30 quantitative radiomic features (e.g., tumor radius, area, perimeter, etc.) into the Support Vector Machine (SVM) that classifies tumor malignancy (Cortes & Vapnik, 1995). In the final step, a multi-class CNN uses the original MRI and output from the previous models to assign the patient to one of 5 diagnostic categories: GBM, Meningioma, Schwannoma, Neurocytoma, and Other (Hussain et al., 2018).

U-Nets are a type of convolutional neural network (CNN) primarily used for image-to-image translation tasks, such as medical image segmentation (Ronneberger, Fischer, & Brox, 2015). The goal of using the 3D full-resolution nnU-Net architecture for the segmentation model is, given a pre-treatment MRI image with a tumor, to generate segmentation masks for the edema, enhancing tumor, and tumor core regions (Figure 3) (Isensee et al., 2021). CNNs are well-suited for image-to-image translation because they can learn spatial hierarchies of features from image data and map one image representation to another of the same spatial dimensions (LeCun, Bengio, & Hinton, 2015). In the case of nnU-Net, the model takes a volumetric MRI scan as input and transforms it into a spatially aligned map of tumor subregions (Isensee et al., 2021). Alternative model architectures such as Generative



Adversarial Networks (GANs) were also considered for this segmentation task due to their quick inference times of under 0.1 seconds per volume (Goodfellow et al., 2014). However, due to the severe class imbalance often present in medical segmentation tasks, as well as the wide variability in data quality introduced purposefully through the real-world resembling augmentation, GANs tend to struggle with producing reliable and clinically accurate outputs (Kazemi et al., 2021). Their adversarial training setup can be highly unstable and prone to mode collapse, especially when faced with rare tumor classes or noisy inputs that are common GBM cases (Salimans et al., 2016).

**Figure 3**

Model Generated Segmentations

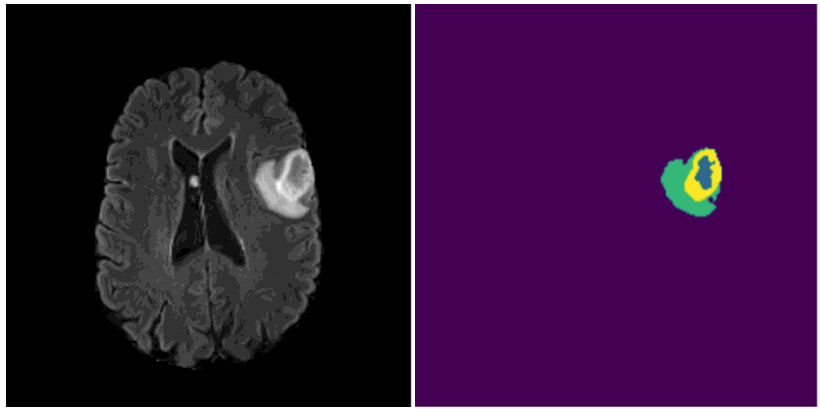

*Segmentations of edema (green), enhanced tumor (yellow), and tumor core (blue)*

To determine the malignancy status, the doctor uses the system-generated segmentation to identify the tumor's 30 quantitative radiomic features (e.g., radius, area, perimeter) (Aerts et al., 2014). The doctor then passes these features into a Support Vector Machine (SVM). The SVM constructs an optimal hyperplane in a high-dimensional feature space that best separates benign from malignant samples (Cortes & Vapnik, 1995). It does so by focusing on the most critical data points near the decision boundary, known as support vectors, and maximizes the margin between the two classes. This margin-based approach not only improves generalization to unseen data but also makes the model resistant to overfitting in high-dimensional contexts (Schölkopf & Smola, 2002). While ensemble-based methods like



Random Forests were considered due to their strength in handling noisy inputs and performing implicit

feature selection (Breiman, 2001), they were ultimately not chosen as they tend to create less distinct

decision boundaries, which reduces clinical interpretability. In contrast, SVMs provide a more

mathematically rigorous and generalizable framework which is of great use for the heterogeneous

nature of GBM (Chang & Lin, 2011).

In addition to CNNs being used for image-to-image translation tasks, they can also be used for

classification purposes (LeCun, Bengio, & Hinton, 2015). CNNs operate by learning spatial hierarchies of

features directly from image data through a series of convolutional operations (Goodfellow, Bengio, &

Courville, 2016). At the core of each convolutional layer is a set of learnable filters, also known as

kernels, typically represented as small matrices (5×5 for binary and 3×3 for multi-class) that slide across

the input image. As each filter moves spatially across the image, it performs an element-wise

multiplication with the underlying pixel values in the local receptive field, producing a feature map that

highlights specific patterns such as edges, textures, or gradients (Krizhevsky, Sutskever, & Hinton, 2012).

Each filter is trained to respond to a different type of feature, and deeper layers capture increasingly

abstract representations. After convolution, a non-linear activation function (Tanh for binary and ReLU

for multi-class) is applied to introduce non-linearity, enabling the network to model complex patterns

necessary in GBM (Nair & Hinton, 2010).

In this system, CNNs are tailored for both binary classification to determine mass presence and

tumor confirmation, and multi-class classification to identify tumor type. For binary tasks, the network

ends with a single neuron activated by a sigmoid function, outputting a probability score, while for

multi-class diagnosis, the final layer uses a SoftMax function across five neurons, each representing a

diagnostic category (GBM, Meningioma, Schwannoma, Neurocytoma, and Other) (Goodfellow et al.,

2016). A chain of CNNs was particularly well-suited as it was more computationally efficient than



monolithic transformer-based architectures due to their localized receptive fields and lower memory

overhead, making them ideal for deployment in real-time or resource-constrained clinical environments

(Vaswani et al., 2017; LeCun et al., 2015).

**Phase 2: Treatment Planning**

The treatment planning stage integrates sequential treatment modeling to personalize therapy

planning for GBM patients based on MRI input (Figure 3). Initially, a diagnosed GBM MRI scan is ingested

into a diffusion-based resection modeling module, which simulates the anatomical changes following

surgical tumor removal (Song et al., 2021). The result is then passed to a transformer-based

radiotherapy simulation that estimates spatial and temporal tumor response, guided by an externally

defined input for the number of weeks of radiotherapy (Dosovitskiy et al., 2020). Following this, a

second diffusion model simulates the effect of chemotherapy on the residual tumor. The MRI generated

from the third model (Chemotherapy Response) is considered the post treatment MRI and is evaluated

by a survival rate calculator, which estimates the patient's overall survival in days (Aerts et al., 2014).

This estimate is then compared against a user-defined survival threshold (e.g., 15%), provided by the

treating physician. If the predicted survival meets or surpasses this threshold, the corresponding

treatment configuration is flagged as optimal. If not, the system iterates using a proximal policy update

system to explore and identify a more effective combination (Schulman et al., 2017).



**Figure 4**

Treatment Planning Phase System Diagram

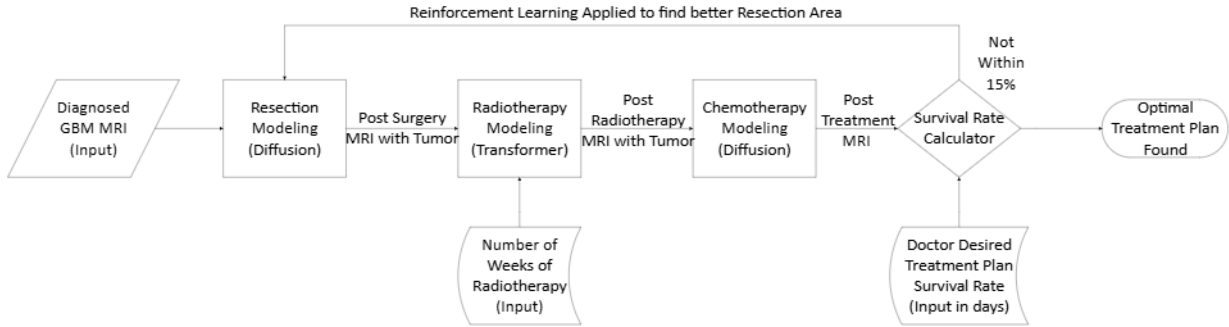

*Flow diagram of the agentic AI treatment planning system for GBM patients. The model sequentially simulates the effects of resection, radiotherapy, and chemotherapy using specialized sub-models before comparing predicted survival outcomes to determine the optimal treatment plan.*

The diffusion model architectures used in predicting resection and chemotherapy outcomes are image-to-image translation models and. Unlike GANs, which suffer from mode collapse and adversarial training instability (Goodfellow et al., 2014), diffusion models produce diverse and realistic outputs by learning a robust denoising process that better captures the complex anatomical features in medical imaging (Ho et al., 2020). In both applications, the model takes an input T1CE MRI image scan of the patient's brain and generates a synthetic post-treatment MRI image that predicts how the brain is expected to appear after the respective intervention. For the resection model, the diffusion model simulates the surgical removal of tumor tissue by learning from examples of pre- and post-resection image pairs (Song et al., 2021). The output is an MRI that reflects changes such as the absence of tumor in surgically accessible regions, potential shifts in brain structure, and other post-surgical features (Figure 5). For the chemotherapy model, the same architecture is used to simulate the biological effects of drug treatment over time. The model is trained on longitudinal imaging data that capture tumor



shrinkage or morphological changes resulting from chemotherapy. The output in this case is a synthetic MRI that visualizes anticipated treatment response, such as partial or complete tumor regression. In both cases, the diffusion model progressively adds noise to the input image during training and then learns to reverse this process during inference, generating realistic and high-resolution images that represent the predicted state of the brain following treatment (Figure 6) (Ho et al., 2020).

**Figure 5**

Resection Models Input Target Pair

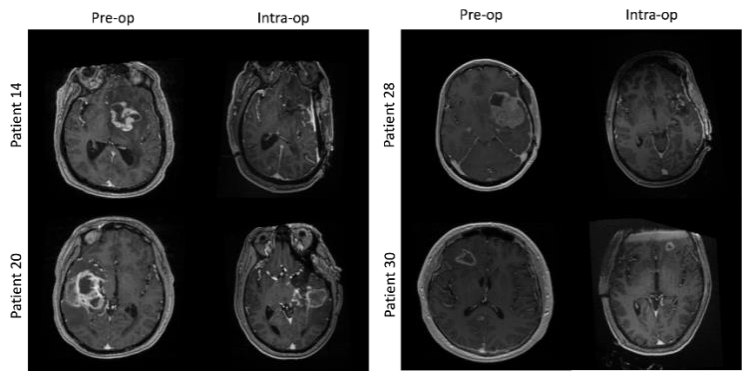

*These images serve as inputs to the training pipeline, enabling the model to learn spatial and structural*

*patterns relevant to tumor identification and surgical planning.*

**Figure 6**

Diffusion Model Architecture Diagram

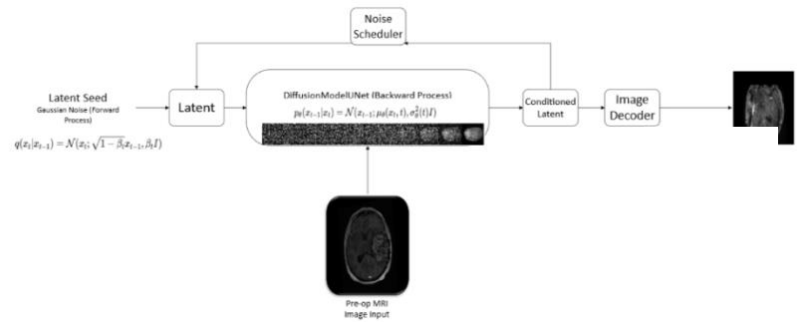

*Diagram showing how a diffusion model learns to generate outputs by denoising random noise,*

*conditioned on both the input and the target label*



The goal of the novel radiotherapy response model (Figure 7) is to analyze intraoperative imaging data and predict brain conditions after *x* weeks of radiotherapy. To achieve this, a Vision Transformer (ViT) based encoder-decoder architecture was designed to model the Spatio-temporal evolution of volumetric brain images over time. The input to the model consists of volumetric data shaped as *(batch, num_weeks, 5, 64, 64, 32)*, representing brain scans collected at multiple time points during treatment. Each volume contains five channels and spatial dimensions of 64×64×32, enabling the model to capture detailed anatomical information across weeks. Recurrent Neural Networks (RNNs), while traditionally used for sequence modeling, were not chosen for this task due to their limited capacity to model long-range dependencies and high-dimensional spatial information inherent in 3D medical imaging data. In contrast, the transformer architecture leverages self-attention mechanisms, which allow the model to attend to relevant spatial and temporal features across all time points simultaneously. This is particularly advantageous for capturing non-linear and global changes in tumor progression that may not follow a strictly sequential pattern. Additionally, transformers support greater parallelization and efficiency during training, making them more scalable for processing large volumetric datasets with multiple temporal slices.



**Figure 7**

Novel Spatio-Temporal Vision Transformer

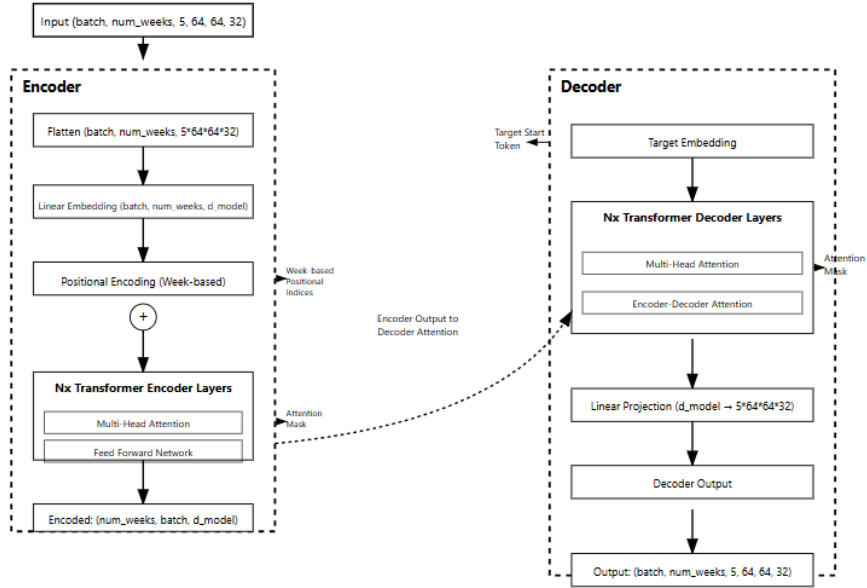

*Transformer-based encoder-decoder architecture for predicting Spatio-temporal brain condition changes*

*from intraoperative radiotherapy imaging data.*

The goal of the novel radiotherapy response model (Figure 7) is to analyze intraoperative imaging data and predict brain conditions after x weeks of radiotherapy. To achieve this, a Vision Transformer (ViT) based encoder-decoder architecture was designed to model the Spatio-temporal evolution of volumetric brain images over time (Dosovitskiy et al., 2020). The input to the model consists of volumetric data shaped as (batch, num_weeks, 5, 64, 64, 32), representing brain scans collected at multiple time points during treatment. Each volume contains five channels and spatial dimensions of 64×64×32, enabling the model to capture detailed anatomical information across weeks. Recurrent Neural Networks (RNNs), while traditionally used for sequence modeling, were not chosen for this task due to their limited capacity to model long-range dependencies and high-dimensional spatial information inherent in 3D medical imaging data (Lipton et al., 2015). In contrast, the transformer architecture leverages self-attention mechanisms, which allow the model to attend to relevant spatial



and temporal features across all time points simultaneously (Vaswani et al., 2017). This is particularly advantageous for capturing non-linear and global changes in tumor progression that may not follow a strictly sequential pattern. Additionally, transformers support greater parallelization and efficiency during training, making them more scalable for processing large volumetric datasets with multiple temporal slices (Child et al., 2019).

The decoder operates autoregressively, generating future brain condition predictions conditioned on previous outputs and the encoded features. It begins with a target start token and embeds subsequent predicted tokens into the same latent space. The decoder layers incorporate self-attention to maintain sequential consistency, along with encoder-decoder attention that grounds predictions in the encoded brain features. Causal attention masks ensure predictions depend only on past and current information, preserving the autoregressive generation. Finally, a linear projection maps the decoder's latent outputs back to the original volumetric shape, producing detailed spatial brain volume predictions. The final output is a tensor representing predicted brain conditions across future weeks, enabling precise monitoring of radiotherapy effects (Vaswani et al., 2017; Child et al., 2019). This Spatio-temporal vision transformer approach introduces several novel contributions to radiotherapy response modeling. By embedding volumetric brain images alongside explicit week-based positional information, the model jointly captures spatial and temporal dynamics critical for accurate prediction. The multi-head attention mechanism enables learning of complex, non-linear temporal relationships in treatment response. The unified encoder-decoder framework supports end-to-end autoregressive forecasting of future brain states based on intraoperative imaging, enhancing predictive performance. Additionally, adaptable attention masks allow flexible sequence handling and enforce temporal causality. Importantly, producing high-resolution volumetric outputs permits detailed, localized predictions essential for clinical decision-making (Dosovitskiy et al., 2020; Vaswani et al., 2017; Parmar et al., 2018).



In the proposed multi-agent reinforcement learning (MARL) framework, the system's feedback is governed by a survival-based reward signal defined as:

$$1 - \left(\frac{|\text{Doctor} - \text{Model}|}{\text{Doctor}}\right) \tag{1}$$

This formulation computes the relative error between the model-predicted survival duration and the physician-defined expected survival target. The reward is maximized when the predicted survival closely matches the clinical goal (Sutton & Barto, 2018). During each training iteration, agents generate treatment actions, which propagate through post-treatment simulation modules (diffusion or transformer-based) (Ho et al., 2020; Vaswani et al., 2017). The resulting post-treatment MRI, along with patient age, is input to a CNN-based controller that predicts survival duration in days. The controller is a custom survival calculator built on a modified 3D ResNet-18 backbone, adapted to accept four imaging channels instead of the standard three (He et al., 2016). After extracting spatial features from the post-treatment MRI, the model concatenates them with the patient's age and passes the combined representation through a multi-layer fully connected regression head. This head is custom designed to capture non-linear relationships between anatomical changes and survival time (Goodfellow et al., 2016). The deviation from the doctor's target is used to compute $R_t$, which serves as the feedback signal for the PPO-based policy update (Schulman et al., 2017).

To ensure clinical relevance and computational efficiency, the system includes a convergence threshold: if the predicted survival falls within ±15% of the physician-defined expectation, the feedback loop is terminated early, and the current treatment plan is accepted as optimal. If the model exceeds the physician's survival goal, the plan is still accepted. Otherwise, the agents continue to update their policies using the PPO update rule (2) and iterates until convergence is met.

$$L^{PPO}(\theta) = \mathbb{E}[\min{(r_t(\theta)\ A_t, \text{clip}(r_t(\theta), 1\text{-}e, 1 + e)\ A_t)}] \tag{2}$$



**Training**

Table 1

Training Configuration of Diagnosis and Treatment Models

| Model | Type | Loss Function | Optimizer | Learning Rate | Epochs | Batch Size |
|---|---|---|---|---|---|---|
| Mass Detection | CNN | Cross Entropy | Adam | 0.0001 | 90 | 32 |
| Tumor Detection | CNN | Cross Entropy | Adam | 0.0001 | 60 | 32 |
| Malignancy Detection | SVM | Hinge | - | - | - | - |
| Diagnosis | CNN | Cross Entropy | SGD | 0.001 | 50 | 30 |
| Segmentation | U-Net | Dice and Cross Entropy | SGD | 0.01 -> 0.0073 | 280 | 2 |
| Resection | Diffusion | MSE | Adam | 2.5e-5 | 800 | 2 |
| Radiotherapy | ViT | MSE | Adam | 1e-4 | 40 | 10 |
| Chemotherapy | Diffusion | MSE | Adam | 1e-4 | 300 | 1 |
| Survival Rate Calculator | CNN | MSE | Adam | 3e-4 | 50 | 2 |

*Overview of the training setup for each model used in both the diagnosis and treatment planning phases. The table details model type, corresponding loss functions, optimization algorithms, learning rates, number of training epochs, and batch sizes*

Training was performed using Google Colab's T4 and A100 GPUs, which offered the high memory and computational bandwidth needed to process 3D MRI scans and large parameter models (Abadi et al., 2016; NVIDIA, 2020). Each model in the system was trained independently with hyperparameters and loss functions specifically tailored to the nature of its task (Table 1). Models were grouped into two phases: the diagnosis phase, which included classification and segmentation models, and the treatment planning phase, which included generative and forecasting models.



In the diagnostic phase, classification models such as the CNNs for mass detection and tumor confirmation utilized categorical cross-entropy loss (3). This loss function is widely used for classification tasks because it penalizes incorrect predictions with greater severity when the model is highly confident, which accelerates convergence (Goodfellow et al., 2016). For the SVM-based malignancy classifier, hinge loss was used as its objective is to maximize the decision boundary between classes, which is particularly effective for separating benign and malignant cases based on radiomic features (Cortes & Vapnik, 1995).

$$L_{CE}(y, p) = -\sum_{c=1}^{C} y_c \log(p_c) \tag{3}$$

$$\mathcal{L}(y, y) = min_{w,b}^2 ||w||^2 + C \sum_{\{i=1\}}^{n} max\left(0, 1 - y_{i(w^\tau x_i + b)}\right) \tag{4}$$

The segmentation model, implemented with nnU-Net, employed a combined Dice loss and cross-entropy loss. Dice loss (5) was chosen due to its effectiveness in medical image segmentation where pixel-wise class imbalance is prevalent; it focuses on overlap between predicted and ground truth segmentations (Milletari et al., 2016). The addition of cross-entropy further improves learning of class boundaries by incorporating pixel-wise probability calibration, making this hybrid loss particularly effective in distinguishing between edema, enhancing tumor, and tumor core (Isensee et al., 2021).

$$\text{Dice Loss} = 1 - \frac{2 * |P \cap G|}{|P| + |G|} \tag{5}$$

All diagnostic models used either Stochastic Gradient Descent (SGD) or Adam optimizer. The CNN classifiers employed the Adam optimizer, which adaptively adjusts learning rates per parameter, allowing for faster convergence and better handling of sparse gradients, which is ideal for early-stage feature detection (Kingma & Ba, 2015). In contrast, SGD was used in the nnU-Net segmentation model because it offers better generalization performance on large-scale image segmentation tasks. It's slower and has more deliberate updates, making it well-suited for fine-grained tasks involving class imbalance, as in tumor segmentation (Ronneberger et al., 2015).



In the treatment planning phase, generative models such as the resection, radiotherapy, and chemotherapy modules used Mean Squared Error (MSE) as the loss function (6). MSE is ideal for tasks involving continuous-valued outputs, as it penalizes the squared difference between predicted and actual pixel intensities. This makes it highly effective in preserving both global anatomical structure and local intensity fidelity in medical imaging (Goodfellow et al., 2016).

$$MSE = \frac{1}{N} \sum_{i=1}^{N} (y_i - \hat{y_i})2 \qquad (6)$$

For example, the resection model was trained for 800 epochs using a learning rate of 1e-4. The use of MSE enabled the model to iteratively refine its predictions of post-surgical brain structure, ensuring that the simulated MRI images closely resembled realistic outcomes (Ho et al., 2020). Similarly, the radiotherapy model was trained for 40 epochs. It relied on MSE to capture subtle changes in tumor volume across multiple time points, stabilizing gradient updates and improving long-term forecasting accuracy (Ronneberger et al., 2015; Dosovitskiy et al., 2020). This consistent use of MSE across treatment models helped the system produce structurally accurate simulations, which were crucial for downstream survival rate prediction and optimal resection decision-making (Chen et al., 2021).

All treatment models used the Adam optimizer, given its effectiveness in optimizing deep generative models with sparse and noisy gradients (Kingma & Ba, 2014). Adam's adaptive learning rates were particularly beneficial during the early epochs of training diffusion and transformer models, where changes in gradient magnitude can be highly volatile. For the diffusion models, which were more computationally intensive and prone to slow convergence, longer training durations (up to 800 epochs) and lower learning rates (1e-4 to 2.5e-5) were applied to stabilize learning without overshooting the loss surface (Ho et al., 2020; Dosovitskiy et al., 2020).



**Results & Discussion**

Table 2

Confusion Matrix for Diagnosis Models

| Classification | TP | TN | FP | FN |
|---|---|---|---|---|
| Mass | 92.1% | 96.7% | 7.9% | 3.3% |
| Tumor | 92.6% | 90.2% | 8.4% | 8.8% |
| Diagnosis | 99.4% | 99.4% | 0.58% | 0.58% |
| Malignant | 97% | 98% | 3% | 2% |

*Performance metrics for diagnosis models, including true/false positives and negatives.*

The classification models developed in this study demonstrated strong performance across all diagnostic stages when evaluated using the confusion matrix. The first-stage binary classifier, which was designed to distinguish between scans with and without masses, achieved a true positive rate (TP) of 92.1% and a true negative rate (TN) of 96.7% (Table 2). These results suggest that the model was effective at identifying abnormal scans while maintaining a low rate of false positives. The false positive rate (FPR) of 7.9%, while moderate, remains acceptable in early diagnostic contexts where sensitivity is prioritized to avoid missing potential tumor cases (Fawcett, 2006). In such settings, it is clinically safer to flag a slightly higher number of normal scans than to risk overlooking abnormal ones. A moderate FPR ensures that nearly all tumors are identified, with the small number of false positives easily resolved by expert review. This tradeoff is a well-established principle in diagnostic medicine (Doi, 2007), especially in critical tasks like early tumor detection, where the consequences of false negatives are significantly more severe than those of false positives. The second-stage classifier, which confirmed the presence of tumors among the abnormal cases identified by the first model, achieved a TP of 92.6% and a TN of 90.2%. These values indicate consistent performance, although the false negative rate (8.8%) highlights a potential limitation in the model's ability to detect all true tumor cases.



The final multi-class classification model, responsible for differentiating between GBM, Schwannoma, Neurocytoma, Meningioma, and Other achieved 99.4% accuracy in both true positive and true negative predictions. It also achieved a statistically low false positive/negative of 0.58%. This high level of precision indicates that once the system narrowed its focus to specific tumor cases, it was able to classify tumor types more accurately. This suggests that the model performed better when dealing with well-defined categories, as opposed to the broader, more ambiguous cases handled by the earlier stages of the system. By narrowing the scope at each stage, the multi-stage design of the classification pipeline not only improved accuracy but also helped reduce error propagation and compute (Litjens et al., 2017; Esteva et al., 2019).

To evaluate the performance of the segmentation model, the Intersection over Union (IoU) metric was used as the primary quantitative measure. IoU (7) is a widely adopted metric in medical image segmentation tasks, as it reflects the degree of overlap between the predicted segmentation and the ground truth (Taha & Hanbury, 2015). It is defined as the ratio of the intersection of the predicted region P and the ground truth region G to their union mathematically expressed as:

$$\text{IoU} = \frac{|P \; \cap \; G|}{|P \; \cup \; G|} \tag{7}$$

|P∩G|| denotes the number of pixels (or voxels, in 3D) that are correctly predicted as tumor (i.e., true positives), and |P∪G| includes all pixels that are either predicted or labeled as tumor. An IoU of 1 indicates perfect overlap, while an IoU of 0 signifies no overlap at all. In this study, the segmentation model achieved a peak IoU of 0.89 on the validation set, demonstrating a high degree of spatial alignment between the predicted tumor masks and the annotated ground truth. This is particularly significant given the severe class imbalance inherent in brain MRI segmentation tasks. Given the complexity and heterogeneity in glioblastoma cases where tumor boundaries are often diffuse and



irregular, an IOU score of 0.89 shows that the model consistently segmented key subregions such as the enhancing core and peritumoral edema with high fidelity (Isensee et al., 2021).

**Figure 8**

Segmentation Model IOU & Train/Validation Loss Graph

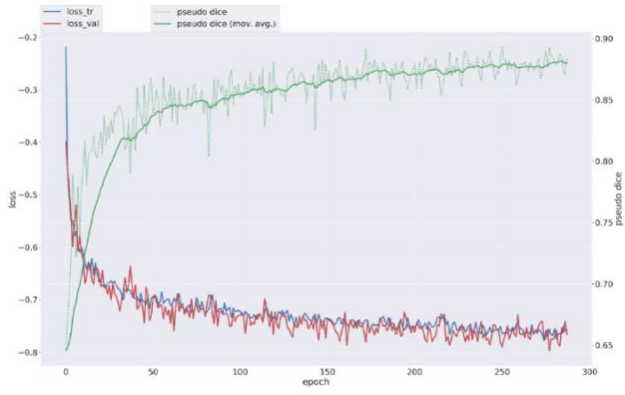

*Segmentation model IOU performance (Max of 0.89) with corresponding train/Val loss trend*

The training and validation loss curves further corroborate the model's performance. As seen in Fig. 8, both the training and validation losses decreased steadily over the course of training, with no significant divergence between the two. This suggests that the model learned generalized features and avoided overfitting, even when trained on a relatively complex dataset. The close tracking of the validation curve to the training curve across epochs indicates stable learning and strong generalization to unseen data. Moreover, the absence of sharp spikes or instability implies that the model converged reliably and was not sensitive to specific samples or noise.

**Treatment Planning Phase**

Each model developed for the treatment planning phase was quantitatively evaluated using the Structural Similarity Index Measure (SSIM) (8), a widely used metric for assessing the perceptual similarity between two images (Wang et al., 2004). SSIM evaluates three key components: luminance,



contrast, and structural similarity between the generated and ground truth images. Mathematically, SSIM is calculated as:

$$SSIM(x, y) = \frac{\{(2\mu_x\mu_y + C_1)(2\sigma_{\{xy\}} + C_2)\}}{\{(\mu_x^2 + \mu_y^2 + C_1)(\sigma_x^2 + \sigma_y^2 + C_2)\}} \tag{8}$$

The SSIM score ranges from −1 to 1, where a value of 1 indicates perfect structural similarity between the predicted and true images. Unlike pixel-wise metrics such as mean squared error (MSE), SSIM is designed to align more closely with human visual perception by focusing on spatial patterns and textures, rather than absolute pixel values alone (Wang et al., 2004). The resection model achieved a maximum SSIM of 0.89, demonstrating that the predicted post-operative MRIs preserved 89% of the structural similarity to actual post-resection scans. This high score indicates that the model effectively learned spatial correspondences between pre-operative tumor regions and realistic surgical outcomes, particularly in anatomically accessible zones. The radiotherapy forecasting model showed a median SSIM of 0.81 across five weekly time points, reflecting its consistent ability to model temporal tumor response dynamics. The interquartile range remained within ±0.03 of the median, highlighting the model's stability and generalization over time. Separately, the chemotherapy response model reached a peak SSIM of 0.82, with a mean squared error (MSE) that stabilized at 0.0131 after 300 epochs. This suggests that the model successfully converged and was capable of reliably simulating drug-induced tumor regression. Overall, the SSIM scores across all models demonstrate their ability to generate structurally accurate and clinically meaningful MRI predictions (Zhao et al., 2017).



Figure 9

Distribution of Radiotherapy Model SSIM Score

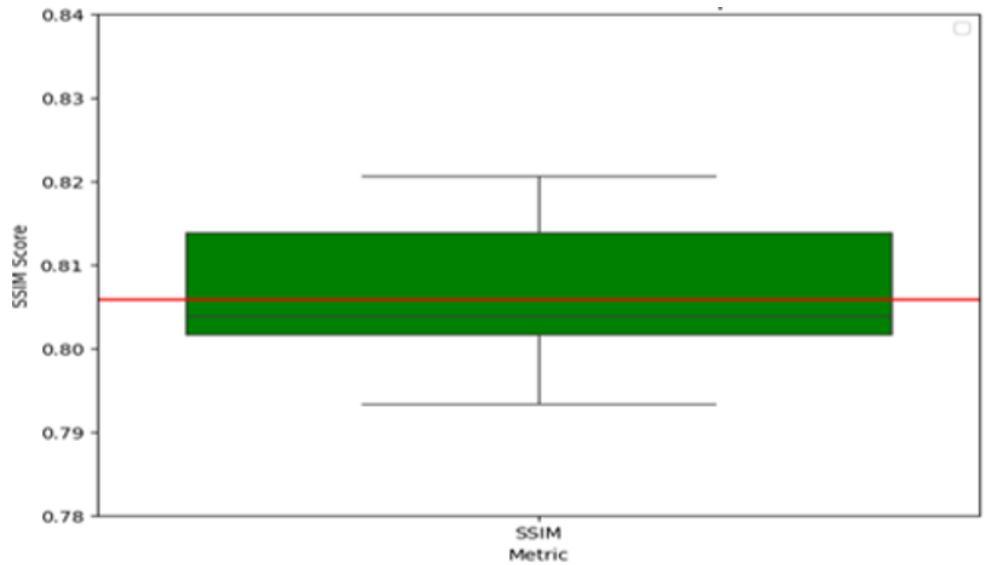

*Box plot showing the distribution of SSIM scores across epochs for radiotherapy analysis*

## Conclusions

This study presents *Brainstorm*, an advanced agentic AI system that is significantly cheaper, faster, and more accurate than current state-of-the-art solutions for GBM diagnosis and treatment planning. Compared to traditional approaches, *Brainstorm* achieved higher segmentation accuracy across all tumor subregions with DICE scores of 0.9095 for edema, 0.8795 for enhancing tumor, and 0.8653 for tumor core (Table 3). This improvement is directly attributed to the use of clinically inspired, real-world augmentations such as bias field distortions, elastic deformations, and random noise transformations which enabled the model to generalize more effectively to the unpredictable anatomical variations seen in actual clinical settings.



Table 3

Segmentation Model Comparison

| Model | Edema | Enhancing Tumor | Tumor Core |
|---|---|---|---|
| **Brainstorm** | **0.9095** | **0.8795** | **0.8653** |
| BraTS 2023 (GAN) | 0.9005 | 0.8673 | 0.8509 |
| BraTS 2020 (U-NET) | 0.8895 | 0.8506 | 0.8203 |

*Evaluation of model performance in segmenting brain tumors into edema, enhancing tumor, and tumor core. Brainstorm outperforms both BraTS 2023 (GAN) and BraTS 2020 (U-Net) across all tumor subregions.*

*Brainstorm* offers a superior alternative to PDE-Based forecasting models with a Spatio-temporal Vision Transformer that utilizes multi head attention to develop faster predictions. Whereas PDE models require up to 225 hours and also forecast tumor growth when no treatment is applied, making it clinically not useful. Brainstorm forecasts the tumors response to all treatments in the treatment cycle, capturing the actual effects of surgery, radiotherapy, and chemotherapy over time. This not only makes Brainstorm's prediction time significantly quicker (10 seconds per forecast), but also makes it clinically viable for use in real-time decision making.

Table 4

Compute Cost Comparison

| Model | Avg. Time For Patient | Time For Simulation |
|---|---|---|
| Brainstorm | 9.7 Seconds | 9.7 Seconds |
| Glioma Solver | 31 Hours | 110.5 Seconds |
| PDE Frameworks | 83 Hours | 300 Seconds |
| Lattice Boltzmann | 225 Hours | 850 Seconds |

*Assessment of average forecasting and simulation times per patient. Brainstorm demonstrates superior speed and efficiency over traditional models and physics-based frameworks.*



A key attribute of this system is the novel sequential decision-making framework which uses a chain of lightweight classifiers to progressively narrow the scope of the possible diseases the patient could have. This allows for a less reliance of one single, large diagnostic model and significantly reduces model complexity and cuts computational costs by 22.28× (Table 5) without compromising performance, making *Brainstorm* scalable and accessible even in lower-resource settings.

Table 5

Segmentation Model Comparison

| Model | Parameters | Cost | Accuracies |
|-------|-----------|------|-----------|
| Brainstorm | 20,891,829 | ~$4,600 | 95.28% |
| R50-ViT-l16 | 23,531,175 | ~$4,600 | 90.31% |
| ViT-b16 | 87,466,855 | ~$26,800 | 97.89% |
| ViT-l16 | 303,317,095 | ~$102,420 | 97.08% |

*Comparison of parameter count, training cost, and accuracy across various models. Brainstorm offers the highest accuracy at a significantly lower cost than larger models like ViT-L16.*

This research was reviewed and affirmed through direct consultations with practicing physicians. These physicians projected a 0.9% increase in five-year survival for GBM patients. This is equivalent to over 2,250 additional lives out of the 250,000 lives being saved annually. By being an all-in-one encompassing system with capabilities of aiding doctors throughout the treatment cycle, it shows a new end to end implementation that allows hospitals to easily switch to. Because of this, the consulted physician also estimated that with rural clinics getting advanced planning tools, it will increase the worldwide patient coverage from 28.7% to 34.9%. Together, these advancements establish *Brainstorm* as a robust, efficient, and clinically grounded system that sets a new benchmark for intelligent, outcome-driven neuro-oncology.



Table 6

Societal Impact

| State | % of Patients | Treatment Cost | 5 Year Survival Rate |
|-------|---------------|----------------|----------------------|
| Brainstorm | 34.9% | $11,993 - $173,641 | 6% (Saves 2250 Lives) |
| Current State | 28.7% | $14,110-$204,284 | 5.1% |

*Projected real-world impact of Brainstorm's integration into treatment planning. Includes cost reduction and improvement in 5-year survival rates compared to current standards.*

**Future Work**

Following in-depth discussions with medical professors at Harvard Medical school and practicing physicians, it has become evident that the architectural design and task structure of *Brainstorm* originally developed for GBM are broadly applicable to other locally invasive cancers such as breast cancer and pancreatic cancer. These cancers, like GBM, require multimodal treatment planning involving surgical resection, radiotherapy, and chemotherapy, making them ideal candidates for *Brainstorm*'s modular treatment-aware architecture. Leveraging this insight, the next phase of this research will involve adapting and validating the system on breast cancer cases. Expanding to multiple cancer types will not only increase the system's impact in real world setting, but also broaden its potential for adoption across a wider spectrum of hospitals and care centers.